\documentclass[10pt,twocolumn,letterpaper]{article}

\usepackage{iccv}
\usepackage{times}
\usepackage{epsfig}
\usepackage{graphicx}
\usepackage{amsmath}
\usepackage{amssymb}

\usepackage{listings}
\usepackage{booktabs}
\usepackage{color}
\usepackage{graphicx}
\usepackage{multirow}
\usepackage{colortbl}

\usepackage{xcolor}

\usepackage{resizegather}

\newcommand\blfootnote[1]{%
  \begingroup
  \renewcommand\thefootnote{}\footnote{#1}%
  \addtocounter{footnote}{-1}%
  \endgroup
}

\usepackage[pagebackref=true,breaklinks=true,letterpaper=true,colorlinks,bookmarks=false]{hyperref}

\iccvfinalcopy 

\def\iccvPaperID{2085} 

\ificcvfinal\fi

\begin{document}

\title{Self-Supervised Monocular Depth Estimation by \\ Direction-aware Cumulative Convolution Network}

\author{
Wencheng Han$^{1}$,
Junbo Yin$^{2}$,
Jianbing Shen$^{1\dagger}$\\
$^1$ SKL-IOTSC, CIS, University of Macau,
$^2$ Beijing Institute of Technology \\
{\tt\small \{wencheng256, yinjunbocn, shenjiangbingcg\}@gmail.com} \\
}

\maketitle
\ificcvfinal\thispagestyle{empty}\fi

\begin{abstract}
Monocular depth estimation is known as an ill-posed task in which objects in a 2D image usually do not contain sufficient information to predict their depth. 
Thus, it acts differently from other tasks (\textit{e.g.}, classification and segmentation) in many ways. 
In this paper, we find that self-supervised monocular depth estimation shows a direction sensitivity and environmental dependency in the feature representation. 
But the current backbones borrowed from other tasks pay less attention to handling different types of environmental information, limiting the overall depth accuracy.
To bridge this gap, we propose a new Direction-aware Cumulative Convolution Network (DaCCN), which improves the depth feature representation in two aspects. First, we propose a direction-aware module, which can learn to adjust the feature extraction in each direction, facilitating the encoding of different types of information. 
{Secondly, we design a new cumulative convolution to improve the efficiency for aggregating important environmental information}. Experiments show that our method achieves significant improvements on three widely used benchmarks, KITTI, Cityscapes, and Make3D, setting a new state-of-the-art performance on the popular benchmarks with all three types of self-supervision. \href{https://github.com/wencheng256/DaCCN}{https://github.com/wencheng256/DaCCN}.
\blfootnote{Corresponding author$^{\dagger}$: \textit{Jianbing Shen}. This work was supported in part by the FDCT grants 0154/2022/A3 and SKL-IOTSC(UM)-2021-2023, the MYRG-CRG2022-00013-IOTSC-ICI grant and the SRG2022-00023-IOTSC grant.
}

%
\end{abstract}

\section{Introduction}
\label{Sec:intro}
Monocular depth estimation is an important vision task for autonomous driving, which can generate a depth map for the image from a single camera. Unlike stereo-matching methods~\cite{rogister2011asynchronous,geiger2010efficient,liu2014optimized,laga2020survey}, monocular depth estimation does not require rectified images, making it easier to be applied for self-driving cars. 
Because of this, monocular depth estimation methods attract much more attention from both the academic and the industrial societies, and many representative monocular depth estimation methods \cite{li2018undeepvo,li2019sequential,chen2019self,monodepth} have been proposed during the last decade.

\begin{figure}[t]
  \centering
  \mbox{}\hfill
  \includegraphics[width = 0.99 \linewidth]{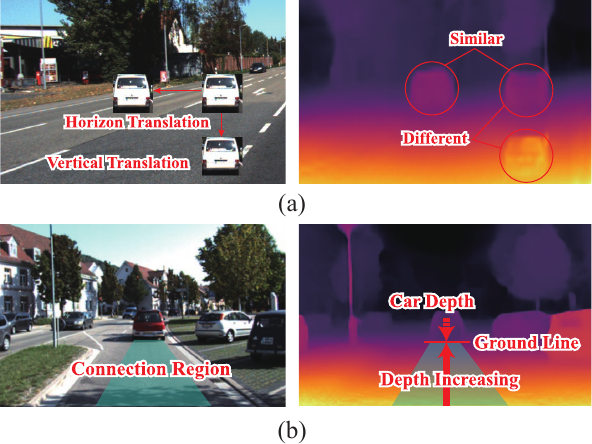}
\caption{\textbf{Illustration of the direction sensitivity of self-supervised monocular depth estimation.} In (a), we translate the same car into different positions, and their depth values are shown in the right figure. In (b), we illustrate the connection region of the car and analyze the importance of this region.
  }
  \label{Fig:motivation}
\vspace{-5mm}
\end{figure} 

The pioneering work of Eigen \textit{et al.}~\cite{eigen_split} first developed a CNN-based network and trained the model in a fully supervised manner. 
To alleviate the need for the ground truth, Grag \textit{et al.}~\cite{monodepth} proposed a self-supervised method based on the stereo images. 
Zhou \textit{et al.}~\cite{zhou2017unsupervised} proposed a pose network to predict the relative position between two consecutive frames and only employ the sequence captured by a single camera in the training phase.
{Based on these works, a series of monocular depth estimation methods based on self-supervised learning have been proposed~\cite{poggi2020uncertainty,watson2019self,kumar2020unrectdepthnet,poggi2020uncertainty}.}
In this paper, we mainly focus on the self-supervised monocular depth estimation task by fully exploring the direction sensitivity and environmental dependency information of this task.

Monocular depth estimation is an ill-posed task since the pixels of one object do not contain enough information to predict its depth. Therefore, the models highly rely on the interrelationships between the objects and environments. 
{Previous depth estimation backbones~\cite{unet,sun2019high,he2016deep,sandler2018mobilenetv2} seldom considered the depth-aware environmental encoding efficiency, which will lead to the lack of important depth clues, thereby limiting the overall performance of models.}
\begin{table}[t]
\small
\centering
\resizebox{0.45\textwidth}{!}
  {
\setlength{\tabcolsep}{4mm}{
\begin{tabular}{cccc}
\bottomrule
\multicolumn{1}{c|}{\multirow{2}{*}{Settings}} & \multicolumn{3}{c}{Metrics}                                                                                   \\ \cline{2-4} 
\multicolumn{1}{c|}{}                             & \multicolumn{1}{c}{Abs Rel $\downarrow$} & \multicolumn{1}{c|}{RMSE $\downarrow$} & $FLOPs$ \\ \hline
\multicolumn{1}{c|}{original ($640 \times 192$)}                        & \multicolumn{1}{c}{0.115}                & \multicolumn{1}{c|}{4.863}             & 8B                     \\ \hline
\multicolumn{1}{c|}{Horizon Stretch ($1280 \times 192$)}                        & \multicolumn{1}{c}{0.118}                & \multicolumn{1}{c|}{4.875}             & 16B                     \\ \hline
\multicolumn{1}{c|}{Vertical Stretch ($640 \times 384$)}                   & \multicolumn{1}{c}{0.108}                & \multicolumn{1}{c|}{4.622}             & 16B                     \\ \hline
\multicolumn{1}{c|}{Equal Stretch ($1280 \times 384$)}                        & \multicolumn{1}{c}{0.109}                & \multicolumn{1}{c|}{4.723}             & 32B                     \\ \hline
\end{tabular}}}
\vspace{1mm}
\caption{\textbf{Analysis about different input ratios with monodepth2.} We adopt Abs Rel, and RMSE as our metrics. For the two metrics, lower values are better. We also provide the FLOPs of each setting.}
\label{Table:ratio}
\vspace{-5mm}
\end{table}

In Fig.~\ref{Fig:motivation}(a), the car is translated to different positions in the image, and their depth values are visualized in the right figure. 
From the visualization results, we find that even with the same pixels, these objects in different positions own different depth values. 
{This demonstrates that depth prediction relies on the environment of objects.}
We further observed that the horizontally translated objects have little depth variance from the original object, but the depth of vertically translated objects changed a lot. 
Based on these observations, we infer that information from different directions plays different roles in depth estimation. 
The information along {the view line} contributes more to the depth variations, and the information from the horizontal lines keeps the depth consistency between objects. Therefore feature extraction from each direction could show different preferences. To explore their differences, we prepare a more detailed analysis in Table~\ref{Table:ratio}.

As mentioned in previous works~\cite{monodepth2, brnet}, increasing the input resolution will facilitate detailed information extraction, and a small resolution is helpful for global information encoding. 
Thus, we change the horizon and vertical resolutions, respectively, and train the model to compare their performances. 
If the feature extraction from the two directions contributes equally to the final accuracy, models with the large horizon and vertical resolution will perform similarly. 
As shown in Table~\ref{Table:ratio}, the depth estimator gets a significant performance drop when increasing the horizon resolution, indicating that the global information is preferred in this direction for better performance.
While the model with a large vertical resolution obviously outperforms the one with the original resolution, which performs closely to the model with equally stretched inputs. 
This demonstrated that detailed information is more critical in the vertical direction for performance. 
Along this direction, we infer that information from the connection region is an important clue for depth estimation. 
As shown in Fig.~\ref{Fig:motivation}(b), the ground line is a crucial reference for the depth estimation of the car~\cite{hoiem2005automatic}, while the depth of the ground line largely relies on the region between it and the camera, which is named the connection region in this paper.


Although the depth estimation task is direction-sensitive and environmentally dependent, current backbones cannot fully use these properties. Traditional convolutional networks usually have the same receptive fields for every direction and encode the information from them similarly. This would lead to less efficiency in extracting various types of features. Moreover, convolutional operations equally aggregate information from the receptive fields into the center position. 
{This aggregating strategy cannot efficiently utilize the critical information encoded in the connection regions.}

To solve these problems, we propose a new Direction-aware Cumulative Convolution Network (DaCCN) for depth feature encoding.
Our DaCCN improves the feature representation in two aspects. The first improvement is for feature extraction. DaCCN can learn to adjust the feature extraction from different directions, facilitating their information encoding. 
As discussed above, the encoded information from different directions in the images plays diverse roles during depth estimation. Therefore, the feature extraction from each direction should be adjusted according to the features it carries. 
%
Instead of manually adjusting the feature extraction, we design a learnable and direction-aware module to optimize it in an end-to-end manner during the offline training. 

Another improvement is for feature aggregation. DaCCN can efficiently aggregate environmental information from the connection regions. The connection regions are the areas that contain all the spaces between the camera and objects and are critical for depth estimation.  
To efficiently aggregate information from these areas, we propose a new cumulative convolution operation, which can accumulate the environmental features from the connection regions and learn to fuse them efficiently.
We integrate our DaCCN into a state-of-the-art baseline model of self-supervised depth estimation and evaluate the performance on three representative benchmarks. Experimental results show that our method achieves significant improvements with a new start-of-the-art performance. 

In conclusion, the main contributions of this paper could be summarized into four folds:
\begin{itemize}
    \item We carefully analyze the direction sensitivity and environmental dependency of self-supervised monocular depth estimation and propose the new Direction-aware Cumulative Network for better feature representation in depth estimation. 
    
    \item We find that features extracted from different directions in the image play distinct roles during depth prediction and propose a learnable module to adjust the sample density and receptive fields for each direction.
    
    \item We propose a new convolutional operation for encoding the critical environmental information from the connection regions.
    
    \item Experiments on three datasets show the improvements of the proposed methods, and we set a new state-of-the-art performance on three widely used benchmarks.
\end{itemize}

\section{Related Works}

\subsection{Supervised Monocular Depth Estimation}
Depth estimation is a fundamental task in the computer vision area. It takes RGB images as input and generates depth maps as output. Each pixel in the depth map indicates the corresponding distance between the object and the camera viewpoint. Depth estimation can be functionally classified into three categories, monocular depth estimation, binocular depth estimation, and multi-view depth estimation. Among them, monocular depth estimation has drawn much attention in recent years~\cite{zhou2022self,hui2022rm,patil2022p3depth,xu2022self,wang2019unos,swami2022you,agarwal2022depthformer,swami2022you}, because of its wide application in autonomous driving. 

The supervised learning approach for monocular depth estimation was first introduced, where pixel-level ground truth depth information is needed in the training phase. 
Eigen \textit{et al.}~\cite{eigen_split} first proposed a deep learning model to predict the depth values under the supervision of ground truth. 
Their network consists of two deep network stacks, one responsible for encoding coarse-depth information and the other for fine-grained depth information. After this, different methods were proposed to improve the performance, like Li \textit{et al.}~\cite{li2015depth} applied conditional random fields into monocular depth estimation. Some other works exploited the geometric relationship in the images. For example, Qi \textit{et al.}~\cite{qi2018geonet} proposed two networks to estimate the depth and surface normal from an image. Ummenhofer \textit{et al.}~\cite{ummenhofer2017demon} developed a network to predict the depth maps according to the structure from motion (SfM) technique. 
Although these works have achieved promising performance, the supervised training needs a large amount of ground truth depth, which can only be gained by some special facilities, like LiDAR. The high costed data collection limits the wide application of these methods.

\subsection{Self-supervised Monocular Depth Estimation}
To avoid the need for labelled data in the monocular depth estimation, Garg \textit{et al.}~\cite{garg2016unsupervised} firstly introduced a promising procedure to learn depth estimation in a self-supervised way. They employed stereo images in the training phase and formed depth optimization to the minimizing of disparity between fabricated images and real images. To release the requirement of stereo images, Zhou \textit{et al.}~\cite{zhou2017unsupervised} estimated depth map and camera pose simultaneously and only used video sequences from a single camera in the training phase. 
With the predicted depth and relative pose between adjacent frames, a fabricated frame can be generated, and the disparity can be calculated between it and the real frame. 
But the occlusion pixels between two frames and the moving objects will significantly influence the performance.
Then, Godard \textit{et al.}~\cite{monodepth2} added a minimum loss for alleviating the crucial challenges using self-supervised approaches. They found that the occlusion in the previous and subsequent frames is complimentary. 
Therefore the model could choose the visible frame to calculate the losses of some areas. 
To solve the moving object problem, they propose an efficient strategy by adding another minimum loss to ignore the loss values from these objects. After that, many works improved the performance of self-supervised monocular depth estimation~\cite{ranjan2019competitive, zou2018df, ranftl2020towards,lee2019big,gordon2019depth}. Masoumian \textit{et al.}~\cite{masoumian2021absolute} developed a multi-scale monocular depth estimation based on a graph convolutional network. Guizilini \textit{et al.}~\cite{guizilini20203d} proposed a 3D packing network in this area. Watson \textit{et al.}~\cite{watson2021temporal} introduced the cost volume to build a multi-frame model and achieved significant improvement. Zhou \textit{et al.}~\cite{zhou_diffnet} exploited the semantic information with down and up-sample procedures to improve depth estimation accuracy. 
However, most of these works employed a backbone network from classification-based tasks~\cite{garcia2017review,lu2007survey}, like U-Net~\cite{unet} and HRNet~\cite{sun2019high}, {but few works considered the difference between depth estimation and their original tasks}.

\section{Method}


\begin{figure*}[t]
  \centering
  \resizebox{\textwidth}{!}{
  \includegraphics{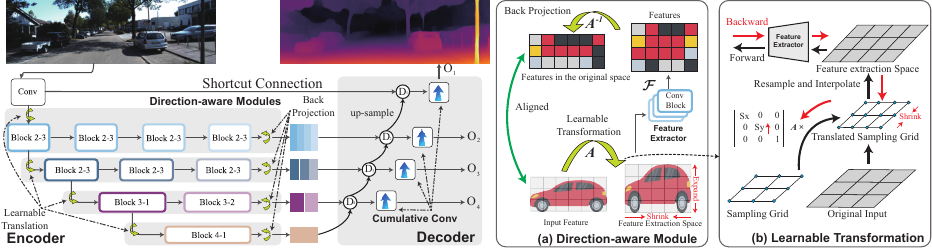}}
  \vspace{1mm}
  \caption{\textbf{The overview architecture of the proposed method.} There are mainly two modules in the proposed DaCCN architecture, a feature encoder, and a depth decoder.  
  (a) shows the details of the direction-aware module. (b) illustrates how the learnable transformation is optimized.
  }
  \label{Fig:architecture}
\vspace{-4mm}
\end{figure*}

\subsection{Direction-aware Cumulative Network}
\label{network}
{Fig.~\ref{Fig:architecture} shows an overview framework of the proposed DaCCN. 
The network includes two main parts: an encoder that extracts feature maps from the input images and a decoder that produces the depth maps based on the feature maps.}
There are four branches in the encoder, each of which encodes features in a different resolution. To achieve the direction-aware feature extraction, we insert a learnable affinity transformation at the beginning of each branch, converting the input into the feature extraction space for direction-aware information encoding. Correspondingly, a back projection is appended at the end of each branch to keep the consistency between features and input images. The affinity transformation, the back projection, and the blocks between them together constitute the direction-aware module. Finally, the outputs of each block in the branches are concatenated and sent into the decoder for depth prediction.

There are four stages in the decoder, where each stage up-samples the current feature maps and fuses them with the corresponding feature maps from the encoder. Finally, a depth map is generated based on the fused feature map. Totally, four stages generate four depth maps with different resolutions, and four losses are calculated with the outputs. In the evaluation phase, only the depth map with the largest resolution is predicted, and the parameters for the other three heads are not used. We then apply a cumulative convolution on the fused feature maps for direction-aware aggregation. Features in this stage have encoded abundant semantic information and local information. Cumulative convolution will aggregate the desired environmental information from the connection regions and improve the depth estimation accuracy. Notably, we only use a single cumulative convolution at each stage because it is enough to aggregate features from the whole connection region. Direction-aware modules and cumulative convolution are two major improvements of DaCCN. In the subsequent sections, we will introduce their details and how they enhance depth prediction.

\subsection{Direction-aware feature extraction}
\label{adjust}
{Convolutional operations in baseline networks usually similarly treat the information from any direction around the object.} 
This is helpful for instance-aware vision tasks because the semantic information from any direction plays a similar role in these tasks. But as mentioned before, self-supervised monocular depth estimation treats the information from each direction differently. This disparity would lead to less efficiency of the model. To alleviate this problem,  we propose a new direction-aware module that can learn to adjust the feature extraction from each direction.

As discussed in Table~\ref{Table:ratio}, the preferred information from each direction differs for the model performance. Based on this, we think the sample density and receptive fields are two direction-aware factors in feature extraction.  The sampling density is defined as the number of feature vectors extracted from a unit area of the input image. Obviously, the larger the sample density, the more detailed information encoded in the feature maps. On the other side, receptive fields control the range of feature extraction, and larger receptive fields would incorporate larger ranges of pixels into each feature vector. Thus the features would pay more attention to the global information as indicated by some previous works~\cite{miangoleh2021boosting,brnet}. 
{The model should use a smaller receptive field and employ more parameters for the direction that needs to be extracted more detailed information.}
On the contrary, large receptive fields are preferred for the global dependency direction; accordingly, the sample density should be reduced for computation efficiency. The direction-aware module is designed based on this assumption and can learn to adjust the sample density and receptive fields during the training phase.

As shown in Fig.~\ref{Fig:architecture}, there are three parts in the direction-aware module, an affinity transformation $\boldsymbol{A}$, a convolutional feature extraction block $\mathcal{F}$, and back projection transformation $\boldsymbol{A^{-1}}$. The affinity transformation is the most important part of the module, which can transform the inputs into the feature extraction space. 
{In this space, the sampling grid of input features is adjusted according to the information in each direction.}
For the direction that needs detailed information, the transformation will increase the sampling numbers and encode more details in the feature map. And for the direction focusing on a global view, the sampling number is reduced for a larger receptive field. Then, the convolutional feature extractor is employed to extract features from the resampled inputs. Finally, the back-projection transformation will transform the features back to the original space:
\begin{equation}
    \boldsymbol{x} = \boldsymbol{A^{-1}} \mathcal{F}(\boldsymbol{A}\boldsymbol{I}),
\end{equation}
where $\boldsymbol{I}$ is the input of the block.

It is hard for humans to determine the suitable way to extract features from each direction. Therefore, we choose to give the learning ability to affinity transformation. To be specific, we regard the scaling ratios in the affinity matrix as two trainable parameters $s_x$ and $s_y$:
\begin{equation}
    \boldsymbol{A} = 
    \begin{bmatrix}
    s_x & 0 & 0\\
    0 & s_y & 0\\
    0 & 0 & 1
    \end{bmatrix},
\end{equation}

Here, we resample and interpolate the inputs into the feature extraction space. Because the interpolation operation will merge pixels in a differentiable way based on their distance to the sampling positions, the sampling points can be optimized by the gradient descent algorithms, as shown in Fig.~\ref{Fig:architecture}(b). Then, features are extracted from the adjusted inputs. As the feature extraction space is not well aligned with the original inputs, we employ a back projection transformation with the inverse matrix $\boldsymbol{A}^{-1}$ for projecting the feature maps back to the original space.

Although some other methods can also adjust feature extraction during the training phase, such as the deformable convolution~\cite{dai2017deformable}, they cannot achieve a similar goal as the direction-aware module. Deformable convolution learns to extract features from different positions by predicting offsets for the convolution kernels. It is designed to efficiently extract features from different-shaped objects, but the sample density cannot be adjusted during this procedure. In contrast, the direction-aware module can change the sample density in each direction and is more suitable for extracting features with different types.

\begin{figure}[t]
  \centering
  \resizebox{0.45\textwidth}{!}{
\includegraphics{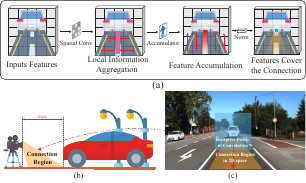}}
  \caption{\textbf{Illustration of the cumulative convolution and connection region.} (a) The cumulative convolution. (b) The connection region in the 3D space. (c) Comparison between the receptive fields of convolutions and cumulative convolution.}
  \label{Fig:cumulation}
\vspace{-5mm}
\end{figure}

\subsection{Cumulative Convolution}
\label{aggregation}
{As discussed in the introduction, monocular depth estimation is a direction-aware task where the information from the connection regions plays the most critical role.}
As shown in Fig.~\ref{Fig:cumulation}(b), in the 3D space, we define the space between the viewpoint and the object as the connection region. 
{It includes the ground between the camera and the object and all the stuff on the ground.}
Therefore, this region contains the most crucial clues for estimating the object's depth value. Given a point $\boldsymbol{P}(X, Y, Z)$ in this region, a corresponding pixel $\boldsymbol{p}(x, y)$ in the 2D image can be gained by applying the intrinsic matrix on it:
\begin{equation}
    \left[ \begin{array}{c}
        x \\
        y \\
        1
        \end{array} 
        \right ]   = \left[ \begin{array}{ccc}
        f_x & 0 & o_x\\
        0 & f_y & o_y\\
        0 & 0 & 1
        \end{array} 
        \right ] \left[ \begin{array}{c}
        X/Z \\
        Y/Z \\
        1
        \end{array}
        \right ],
\end{equation} where $f_x$ and $f_y$ are the pixel focal lengths and $o_x, o_y$ are the offsets of the principal point. 
{Because all the $Z$ values (depth values) in the connection region are smaller than that of the objects, most projected points in the 2D image have larger $y$ values than the objects and thus are located at the bottom areas of the objects.}

We believe that better feature extraction for monocular depth estimation should fully exploit the information from connection regions. However, convolutional operation aggregates information into the center position, covering a square region around the object, making it less efficient to extract the critical information. As shown in Fig.~\ref{Fig:cumulation}(c), the blue areas are the square receptive field of a convolutional operation, and the yellow area indicates the projected connection region. To address this issue, we introduce a new cumulative convolution operation into this task. 
{Instead of simply increasing the receptive fields of the convolution to cover the connection regions, we change the feature aggregation according to the direction where the connection regions are located. }

As shown in Fig.~\ref{Fig:cumulation}(a), there are three parts of this operation. 
The first is a spatial convolution $f$ that can extract spatial information from the local areas and modulate the features for the next stage. The second one is the accumulator $Acc$, which can cumulate the features from the bottom to the current pixels of the feature map. This operation will enlarge the receptive fields of the pixels towards the bottom line covering the whole connection region. But this operation would lead to a value imbalance between pixels because each pixel in the feature map aggregates information from a different scope. 
Therefore, the last part of this operation is a normalization $Norm$ operation which can normalize the aggregated features according to their position in the feature map:
\begin{equation}
    CumulativeConv(\boldsymbol{x}) = \delta(Norm(ACC(f(\boldsymbol{x}, \eta)))), 
\end{equation}
where $\boldsymbol{x}$ is the input features, $\eta$ is the weight in the spatial encoder and $\delta$ is the activation function.

In this paper, we employ a cumulative summation as our accumulator. It will cumulatively sum the features from to bottom to the current positions:
\begin{equation}
    \boldsymbol{X}_{p,q}  = \sum\limits_{i=rows}^{p} \boldsymbol{x}'_{i, q},
\end{equation}
where $p$, $q$ are the row and column index of the resulting feature maps $\boldsymbol{X}$ and $\boldsymbol{x}'_{i, q}$ means features in the $i$th raw and $q$th column of the input feature maps $\boldsymbol{x}'$. $rows$ is the number of rows in $\boldsymbol{x}'$. Correspondingly, we design a normalization method to match this accumulator by dividing the pixels according to their row number:
\begin{equation}
    Norm(\boldsymbol{X})= \frac{\boldsymbol{X}_{p, q}}{(rows -  p)}.
\end{equation}


\subsection{Loss Function}
\label{loss}
%
{Following our baselines, we employ the self-supervised method and formulate the monocular depth estimation problem as minimising the photometric reprojection error.}
To be specific, given two images $I_t$ and $I_{t'}$ from different viewpoints. A pseudo target image $I_{t'\rightarrow t}$is produced by translating the image $I_t'$ according to the predicted depth $D_t$, the relative position $T_{t\rightarrow t'}$ and the intrinsic $K$:
\begin{equation}
I_{t^{\prime} \rightarrow t}=I_{t^{\prime}}\left\langle\operatorname{proj}\left(D_{t}, T_{t \rightarrow t^{\prime}}, K\right)\right\rangle \nonumber
\end{equation}
where stereo images are available, and $T_{t \rightarrow t^{\prime}}$ is the relative position between two cameras; otherwise, it is the predicted position by the PoseNet introduced in~\cite{zhou2017unsupervised}. Then, the disparity $L_p$ between the pseudo image $I_{t'\rightarrow t}$ and the original target image $I_t$ is used to measure the accuracy of the depth $D_t$:
\begin{equation}
L_p=\frac{\alpha}{2}\left(1-\operatorname{SSIM}\left(I_{t'\rightarrow t}, I_t\right)\right)+(1-\alpha)\left\|I_{t'\rightarrow t}-I_t\right\|_{1}, \nonumber
\end{equation}
where two similarity methods are employed to calculate the difference. One is an L1 loss, and the other is the structural similarity loss (SSIM)~\cite{hore2010image}. $\alpha$ is a hyperparameter that controls the weight of the two similarity metrics. Besides, an edge-aware smoothness regulation is used to keep the inner-object disparity smooth:
\begin{equation}
L_{s}=\left|\partial_{x} d_{t}^{*}\right| e^{-\left|\partial_{x} I_{t}\right|}+\left|\partial_{y} d_{t}^{*}\right| e^{-\left|\partial_{y} I_{t}\right|}. \nonumber
\end{equation}

The final loss of our method is defined as: 
\begin{equation}
L= L_{p}+\lambda L_{s}, \nonumber
\end{equation}
where  $\lambda$ is the weight of edge-aware smoothness regulation. 

\section{Experiment}
We train and evaluate our models on a DGX system with an Intel E5-2698 v4CPU, 512G memory. All the training and evaluations are conducted on a single Nvidia V100 GPU.
%
%
To show the improvements, we incorporate them with one newly proposed high-performance baseline DIFFNet~\cite{zhou_diffnet}, {which is based on the HR-Net networks ~\cite{sun2019high,hrdepth} and it is one of the current SOTA works.}

\subsection{Comparison on KITTI}
The KITTI dataset is known as one of the most commonly used vision datasets containing many challenges, such as optical flow~\cite{zhai2021optical}, visual odometry~\cite{nister2004visual}, semantic segmentation~\cite{garcia2017review}. Also, it is considered the most prevalent criterion in self-supervised monocular depth estimation. There are 56 different scenes in the dataset that are divided into 28 scenes for training and the rest for evaluation. We adopt the data split~\cite{eigen_split} as our baseline models and pre-process them as \cite{zhou2017unsupervised} for removing static frames. Finally, $39,810$ triplets are used for training and $4,424$ for validation. 

\begin{table*}[t]
\small
\centering
  \resizebox{\textwidth}{!}
  {
\setlength{\tabcolsep}{3mm}{
\begin{tabular}{l|c|c|c|c|c|c|c|c|c}
\bottomrule
{{Method}} &
{{Resolution}} & {{Trian}} &                {\cellcolor{red!30}Abs Rel} & {\cellcolor{red!30}Sq Rel} & {\cellcolor{red!30}RMSE}  & {\cellcolor{red!30}RMSE log} & {\cellcolor{cyan!20}$\delta < 1.25$} & {\cellcolor{cyan!20}$\delta < 1.25^{2}$} & \cellcolor{cyan!20} $\delta < 1.25^3$ \\ \hline

\hline
\hline
{Monodepth2~\cite{monodepth2}}	&	{$640 \times 192$}	&	{M}	&	{0.115}	&	{0.903}	&	{4.863}	&	{0.193}	&	{0.877}	&	{0.959}	&	\multicolumn{1}{c}{0.981} \\
{PackNet-SfM~\cite{guizilini20203d}}	&	{$640 \times 192$}	&	{M}	&	{0.111}	&	{{0.785}}	&	{{4.601}}	&	{0.189}	&	{0.878}	&	{0.960}	&	\multicolumn{1}{c}{0.982} \\
{HR-Depth~\cite{hrdepth}}	&	{$640 \times 192$}	&	{M}	&	{0.109}	&	{0.792}	&	{4.632}	&	{{0.185}}	&	{0.884}	&	{{0.962}}	&	\multicolumn{1}{c}{{0.983}} \\
{R-MSFM6~\cite{RMSFM}}	&	{$640 \times 192$}	&	{M}	&	{0.112}	&	{0.806}	&	{4.704}	&	{0.191}	&	{0.878}	&	{0.960}	&	\multicolumn{1}{c}{0.981} \\
{DIFFNet~\cite{zhou_diffnet}}	&	{$640 \times 192$}	&	{M}	&	\underline{0.102}	&	{0.764}	&	{4.483}	&	{0.180}	&	\underline{0.896}	&	\underline{0.965}	&	\multicolumn{1}{c}{0.983} \\
{BRNet~\cite{brnet}}	&	{$640 \times 192$}	&	{M}	&	{{0.105}}	&	{\underline{0.698}}	&	{\underline{4.462}}	&	{\underline{0.179}}	&	{{0.890}}	&	{\underline{0.965}}	&	\multicolumn{1}{c}{\underline{0.984}} \\
\rowcolor{green!10}{DaCCN~(ours)}	&	{$640 \times 192$}	&	{M}	&	\textbf{{0.099}}	&	\textbf{{0.661}}	&	\textbf{{4.316}}	&	\textbf{{0.173}}	&	\textbf{{0.897}}	&	\textbf{{0.967}}	&	\multicolumn{1}{c}{\textbf{0.985}} \\
\hline
{Monodepth R50~\cite{monodepth}}	&	{$512 \times 256$}	&	{S}	&	{0.133}	&	{1.142}	&	{5.533}	&	{0.230}	&	{0.830}	&	{0.936}	&	\multicolumn{1}{c}{0.970} \\
{3Net (VGG)~\cite{poggi2018learning}}	&	{$512 \times 256$}	&	{S}	&	{0.119}	&	{1.201}	&	{5.888}	&	{0.208}	&	{0.844}	&	{0.941}	&	\multicolumn{1}{c}{\underline{0.978}} \\
{Monodepth2~\cite{monodepth2}}	&	{$640 \times 192$}	&	{S}	&	{{0.109}}	&	{{0.873}}	&	{{4.960}}	&	{{0.209}}	&	{{0.864}}	&	{{0.948}}	&	\multicolumn{1}{c}{0.975} \\
{BRNet~\cite{brnet}}	&	{$640 \times 192$}	&	{S}	&	{\underline{0.103}}	&	{\underline{0.792}}	&	{\underline{4.716}}	&	{\underline{0.197}}	&	{\underline{0.876}}	&	{\underline{0.954}}	&	\multicolumn{1}{c}{\underline{0.978}} \\
\rowcolor{green!10}{DaCCN~(ours)}	&	{$640 \times 192$}	&	{S}	&	{\textbf{0.099}}	&	{\textbf{0.735}}	&	{\textbf{4.610}}	&	{\textbf{0.193}}	&	{\textbf{0.882}}	&	{\textbf{0.955}}	&	\multicolumn{1}{c}{\textbf{0.979}} \\
\hline
{Monodepth2~\cite{monodepth2}}	&	{$640 \times 192$}	&	{MS}	&	{{0.106}}	&	{0.818}	&	{4.750}	&	{0.196}	&	{0.874}	&	{0.957}	&	\multicolumn{1}{c}{0.979} \\
{HR-Depth~\cite{hrdepth}}	&	{$640 \times 192$}	&	{MS}	&	{0.107}	&	{{0.785}}	&	{{4.612}}	&	{{0.185}}	&	{{0.887}}	&	{{0.962}}	&	\multicolumn{1}{c}{{0.982}} \\
{R-MSFM6~\cite{RMSFM}}	&	{$640 \times 192$}	&	{MS}	&	{0.111}	&	{0.787}	&	{4.625}	&	{0.189}	&	{0.882}	&	{0.961}	&	\multicolumn{1}{c}{0.981} \\
{DIFFNet~\cite{zhou_diffnet}}	&	{$640 \times 192$}	&	{MS}	&	{0.101}	&	{0.749}	&	\underline{4.445}	&	\underline{0.179}	&	\underline{0.898}	&	\underline{0.965}	&	\multicolumn{1}{c}{\underline{0.983}} \\
{BRNet~\cite{brnet})}	&	{$640 \times 192$}	&	{MS}	&	{\underline{0.099}}	&	{\underline{0.685}}	&	{{4.453}}	&	{{0.183}}	&	{{0.885}}	&	{{0.962}}	&	\multicolumn{1}{c}{\underline{0.983}} \\
\rowcolor{green!10}{DaCCN~(ours)}	&	{$640 \times 192$}	&	{MS}	&	{\textbf{0.097}}	&	{\textbf{0.647}}	&	{\textbf{4.282}}	&	{\textbf{0.172}}	&	{\textbf{0.901}}	&	{\textbf{0.967}}	&	\multicolumn{1}{c}{\textbf{0.985}} \\
\hline
\hline
{Monodepth2~\cite{monodepth2}}	&	{$1024 \times 320$}	&	{M}	&	{0.115}	&	{0.882}	&	{4.701}	&	{0.190}	&	{0.879}	&	{0.961}	&	\multicolumn{1}{c}{0.982} \\
{PackNet-SfM~\cite{guizilini20203d}}	&	{$1280 \times 384$}	&	{M}	&	{0.107}	&	{0.802}	&	{4.538}	&	{0.186}	&	{{0.889}}	&	{0.962}	&	\multicolumn{1}{c}{0.981} \\
{HR-Depth~\cite{hrdepth}}	&	{$1024 \times 320$}	&	{M}	&	{{0.106}}	&	{0.755}	&	{4.472}	&	{{0.181}}	&	{{0.892}}	&	{{0.966}}	&	\multicolumn{1}{c}{{0.984}} \\
{R-MSFM6~\cite{RMSFM}}	&	{$1024 \times 320$}	&	{M}	&	{0.108}	&	{0.748}	&	{{4.470}}	&	{0.185}	&	{{0.889}}	&	{0.963}	&	\multicolumn{1}{c}{0.982} \\
{DIFFNet~\cite{zhou_diffnet}}	&	{$1024 \times 320$}	&	{M}	&	\underline{0.097}	&	{0.722}	&	{{4.435}}	&	{\underline{0.174}}	&	{\underline{0.907}}	&	\underline{0.967}	&	\multicolumn{1}{c}{{0.984}} \\
{BRNet~\cite{brnet}}	&	{$1024 \times 320$}	&	{M}	&	{{0.103}}	&	{\underline{0.684}}	&	{\underline{4.385}}	&	{{0.175}}	&	{{0.889}}	&	{{0.965}}	&	\multicolumn{1}{c}{\textbf{0.985}} \\
\rowcolor{green!10}{DaCCN~(ours)}	&	{$1024 \times 320$}	&	{M}	&	{\textbf{0.094}}	&	{\textbf{0.624}}	&	{\textbf{4.145}}	&	{\textbf{0.169}}	&	{\textbf{0.909}}	&	{\textbf{0.970}}	&	\multicolumn{1}{c}{\textbf{0.985}} \\
\hline
{SuperDepth + pp~\cite{pillai2019superdepth}}	&	{$1024 \times 382$}	&	{S}	&	{0.112}	&	{0.875}	&	{4.958}	&	{0.207}	&	{0.852}	&	{0.947}	&	\multicolumn{1}{c}{{0.977}} \\
{Monodepth2~\cite{monodepth2}}	&	{$1024 \times 320$}	&	{S}	&	{{0.107}}	&	{{0.849}}	&	{{4.764}}	&	{{0.201}}	&	{{0.874}}	&	{{0.953}}	&	\multicolumn{1}{c}{{0.977}} \\
{BRNet~\cite{brnet}}	&	{$1024 \times 320$}	&	{S}	&	{\underline{0.097}}	&	{\underline{0.729}}	&	{\underline{4.510}}	&	{\underline{0.191}}	&	{\underline{0.886}}	&	{\textbf{0.958}}	&	\multicolumn{1}{c}{\textbf{0.979}} \\
\rowcolor{green!10}{DaCCN~(ours)}	&	{$1024 \times 320$}	&	{S}	&	{\textbf{0.093}}	&	{\textbf{0.699}}	&	{\textbf{4.450}}	&	{\textbf{0.190}}	&	{\textbf{0.889}}	&	{\underline{0.957}}	&	\multicolumn{1}{c}{\underline{0.978}} \\
\hline
{Monodepth2~\cite{monodepth2}}	&	{$1024 \times 320$}	&	{MS}	&	{0.106}	&	{0.806}	&	{4.630}	&	{0.193}	&	{0.876}	&	{0.958}	&	\multicolumn{1}{c}{0.980} \\
{HR-Depth~\cite{hrdepth}}	&	{$1024 \times 320$}	&	{MS}	&	{{0.101}}	&	{{0.716}}	&	{{4.395}}	&	{{0.179}}	&	{{0.899}}	&	{{0.966}}	&	\multicolumn{1}{c}{{0.983}} \\
{R-MSFM6~\cite{RMSFM}}	&	{$1024 \times 320$}	&	{MS}	&	{0.108}	&	{0.753}	&	{4.469}	&	{0.185}	&	{{0.888}}	&	{0.963}	&	\multicolumn{1}{c}{0.982} \\
{BRNet~\cite{brnet}}	&	{$1024 \times 320$}	&	{MS}	&	{{0.097}}	&	{{0.677}}	&	{{4.378}}	&	{{0.179}}	&	{{0.888}}	&	{{0.965}}	&	\multicolumn{1}{c}{\underline{0.984}} \\
{DIFFNet~\cite{zhou_diffnet}}	&	{$1024 \times 320$}	&	{MS}	&	{\underline{0.094}}	&	{\underline{0.678}}	&	{\underline{4.250}}	&	\underline{{0.172}}	&	{\underline{0.911}}	&	{\underline{0.968}}	&	\multicolumn{1}{c}{\underline{0.984}} \\
\rowcolor{green!10}{DaCCN~(ours)}	&	{$1024 \times 320$}	&	{MS}	&	{\textbf{0.091}}	&	{\textbf{0.622}}	&	{\textbf{4.170}}	&	{\textbf{0.168}}	&	{\textbf{0.912}}	&	{\textbf{0.969}}	&	\multicolumn{1}{c}{\textbf{0.985}} \\
\hline
\end{tabular}}}
\vspace{1mm}
\caption{\textbf{The SOTA comparison on KITTI benchmark-Eigen Split~\cite{eigen_split}}. We compare the proposed methods with the representative models on the KITTI benchmark with three self-supervision manners. \textbf{M} in the train column means training with monocular video sequences, and \textbf{S} means stereo image pairs and \textbf{MS} means training with the two types of data. 
For the {\colorbox{red!30}{error-based metrics}}, the lower value is better; and for the {\colorbox{cyan!20}{accuracy-based metrics}}, the higher value is better.
The best and second best results of each set are marked in \textbf{bold} and \underline{underline}.
}
\label{Table:sota}
\vspace{-4mm}
\end{table*}

\noindent\textbf{SOTA comparison} As shown in Table~\ref{Table:sota}, we evaluate the performance of our DaCCN on Eigen split~\cite{eigen_split}. We roughly divide the results into two types of resolution, \ie the low resolution and the high resolution in the table. Totally we employ 7 metrics in the comparison where $Abs Rel$, $Sq Rel$, $RMSE$, $RMSE log$ are error-based metrics and therefore lower values are better. $\delta$ is the disparity between the predicted depth and ground truth values. $\delta < 1.25$, $\delta < 1.25^2$, $\delta < 1.25^3$ are accuracy-based metrics, and the higher values are better. According to the table, our DaCCN achieves the best performance on all three supervision types and two different input resolutions. 

{Obviously, the improvements on $Sq Rel$ and $RMSE$ are particularly prominent.}
The two metrics are based on square error and the large error values from the hard cases that can be magnified in these two metrics:
\begin{equation}
Sq Rel = \frac{1}{n} \sum\left(\frac{y_{p r e d}-y_{g t}}{y_{g t}}\right)^{2} \nonumber
\end{equation}
\begin{equation}
RMSE = \sqrt{\frac{1}{n} \sum\left(y_{p r e d}-y_{g t}\right)^{2}} \nonumber
\end{equation}
Therefore, improvements in these two metrics indicate that our methods can fix some hard cases in the original model. 
Compared to our baseline model DIFFNet~\cite{zhou_diffnet}, our DaCCN with monocular video training and small inputs achieves 0.003, 0.103 and 0.167 improvements in terms of $AbsRel$, $Sq Rel$ and $RMSE$, respectively.  

With MS training, the improvement of the proposed methods is more significant. With $640 \times 192$ inputs, DaCCN achieves 0.004, 0.102 improvements in terms of AbsRel and RMSE. 
With $1024 \times 320$ inputs and MS training,
our DaCCN also achieves the best results among all these methods and sets a new state-of-the-art performance. 

\begin{figure*}[t]
\centering
\resizebox{0.98\textwidth}{!}
  {
  \includegraphics[width = 15cm]{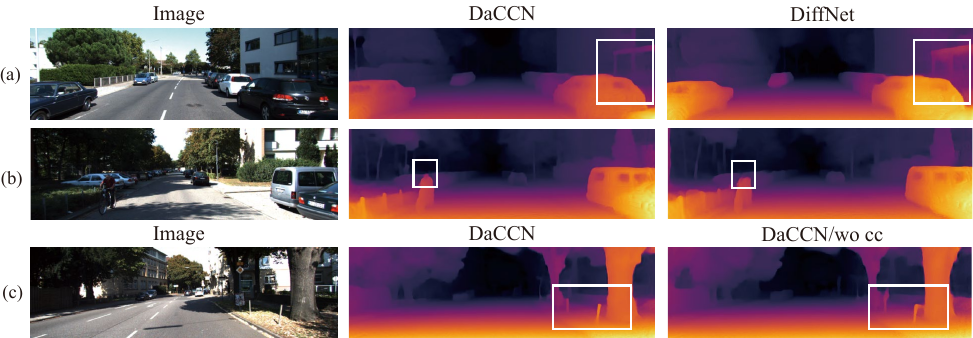}}
  \caption{\textbf{Qualitative results on the KITTI Eigen split test set.} Our DaCCN can correct some errors of previous works (marked in white boxes) owning to the better feature representation.}
  \label{Fig:qualitive}
\vspace{-2mm}
\end{figure*}

\noindent\textbf{Quantitive Results} We also compare the qualitative performance with the baseline work DiffNet~\cite{zhou_diffnet}. As shown in Fig.~\ref{Fig:qualitive}, for some hard cases, our DaCCN can provide the correct depth estimation result while DiffNet cannot. Also, we show a typical improvement case of the cumulation convolution module (CC). In this case, CC can easily correct some vertical errors in the prediction, as shown in the white box in (c). 

\subsection{Comparison on Make3D and Cityscapes}
\noindent\textbf{Make3D} is a dataset including both monocular RGB images and its corresponding depth maps. 
Due to the non-existence of stereo images and monocular sequences, this dataset cannot be used to train self-supervised monocular depth estimation models but is extensively used as a testing set to evaluate the capability of networks on a disparate dataset. 
We compare our models with other representative works on this benchmark. As shown in Table~\ref{Table:make3d}, our models outperform all the other methods, which demonstrates our models can be well generalized to unseen scenes.
With monocular training and $640 \times 192$ input, our method achieves $0.290$ and $6.656$ in terms of $Abs Rel$ and $RMSE$ with significant improvements from other SOTA models.

\noindent\textbf{Citiscapes} is a representative dataset in the semantic segmentation area for autonomous driving applications. Also, it includes a series of stereo video sequences that can be used to train self-supervised depth estimation models. Following the setting of~\cite{watson2021temporal}, we train and evaluate the proposed DaCCN on the Cityscape dataset, and the results are shown in Table~\ref{Table:city}. Our DaCCN significantly outperforms many state-of-the-art models on this dataset.

\begin{table}[t]
\centering
  \resizebox{0.45\textwidth}{!}
  {
\setlength{\tabcolsep}{4mm}{
\begin{tabular}{c|c|c|c|c}
\hline
\multicolumn{2}{c|}{Components} & \multicolumn{1}{c|}{\multirow{2}{*}{Abs Rel $\downarrow$}} & \multicolumn{1}{c|}{\multirow{2}{*}{RMSE $\downarrow$}} & \multirow{2}{*}{$\delta < 1.25$ $\uparrow$} \\ \cline{1-2}
\multicolumn{1}{c|}{DaM} & \multicolumn{1}{c|}{CC} & \multicolumn{1}{c|}{} & \multicolumn{1}{c|}{} & \\ \hline\hline
\multicolumn{1}{l|}{} & \multicolumn{1}{l|}{} & 0.102 & 4.483 & 0.896 \\ \hline
$\checkmark$  & & 0.100 & 4.372 & 0.899 \\ \hline
& $\checkmark$  & 0.101 & 4.380 & 0.897 \\ \hline
$\checkmark$ & $\checkmark$ & 0.099 & 4.316 & 0.897 \\ 
\hline

\end{tabular}}}
\vspace{1mm}
\caption{\textbf{Ablation study of the proposed DaCCN.} We conduct ablation studies on the two new components in our model. Our DaM means the direction-aware module and CC denotes the cumulative convolution operation.
}
\label{Table:ablation}
\vspace{-2mm}
\end{table}

\begin{table}[t!]
\footnotesize
\centering
\resizebox{0.45\textwidth}{!}{
\setlength{\tabcolsep}{2mm}{
\begin{tabular}{c|c|c|c|c}
\hline
Architecture & Abs Rel $\downarrow$ & Sq Rel $\downarrow$ & RMSE$\downarrow$ & {$\delta < 1.25$ $\uparrow$} \\
\hline
\hline
Struct2Depth~\cite{casser2019unsupervised}    & 0.145                & 1.737             & 7.280 & 0.813 \\
\hline
Monodepth2~\cite{monodepth2}    & 0.129                & 1.569            & 6.876 & 0.849 \\
\hline
Videos in the Wild~\cite{gordon2019depth}   & 0.127                & 1.330            & 6.960 & 0.830 \\
\hline
Li et al.~\cite{li2021unsupervised}    & 0.119                & \textbf{1.290}            & 6.980 & 0.846 \\
\hline
\rowcolor{green!10} DaCCN   & \textbf{0.113}                & {1.380}             & \textbf{6.305} & \textbf{0.888} \\
\hline
\end{tabular}
}}
\vspace{1mm}
\caption{\color{black}\textbf{Cityscape results follow the settings of~\cite{watson2021temporal}.}
}
\label{Table:city}
\vspace{-2mm}
\end{table}

\subsection{Ablation Study}
We conduct several ablation studies on the KITTI dataset. 
{Eigen split~\cite{eigen_split} is used to validate the effectiveness of the proposed modules: direction-aware module and cumulative convolution module.}
In this experiment, we employ $Abs Rel$, $RMSE$, and $\delta < 1.25$ as the metrics.

\noindent\textbf{Direction-aware Module (DaM)} Direction-aware module is responsible for adjusting feature extraction from different directions. During the training phase, $s_x$ and $s_y$ are optimized by the gradient descent algorithm to obtain an optimal solution. In our experiment, the optimal $s_y$ is usually larger than $s_x$, because the vertical direction encodes more relative depth information, and the model needs more details to exploit it fully.
On the contrary, the information extracted from the horizon direction reveals the consistency of both inner and inter objects.
Thus, the model needs larger receptive fields to achieve this. As shown in Table~\ref{Table:ablation}, our DaM improves the performance on all the metrics. 

\noindent\textbf{Cumulative Convolution (CC)} is responsible for enhancing environmental information aggregation. As shown in Table~\ref{Table:ablation}, the improvements of this module mainly lie in the $RMSE$ metric, which means this module corrects some hard cases for depth estimation. Environmental information is critical for the depth prediction of objects, while the original CNN cannot efficiently aggregate this information from the connection regions and thus fail to estimate depth values for some targets. 
{Our cumulative convolution compensates for this limitation and improves the overall performance.}

\noindent\textbf{Efficiency} 
We also compare the efficiency with more state-of-the-art approaches. In the training phase, the computation of our model will change because the sampling density is optimized by the gradient-descendent algorithm. 
{In the evaluation phase, the sampling density has reached its optimal solution and thus the computation is fixed.}
As shown in Table~\ref{Table:flops}, compared with other well-known methods, our model achieves a good balance between performance and efficiency. 

\begin{table}[t!]
\footnotesize
\centering
\resizebox{0.45\textwidth}{!}{
\setlength{\tabcolsep}{3mm}{
\begin{tabular}{c|c|c|c|c}
\hline
Architecture & Abs Rel $\downarrow$ & Sq Rel $\downarrow$ & RMSE$\downarrow$ & $log_{10}$  $\downarrow$ \\
\hline
\hline
Monodepth    & 0.544                & 10.94             & 11.760 & 0.193 \\
\hline
Monodepth2    & 0.322                & 3.589             & 7.414 & 0.163 \\
\hline
BRNet    & {0.302}                & {3.133}             & {7.068} & {0.156} \\
\hline
\rowcolor{green!10} DaCCN    & \textbf{0.290}                & \textbf{2.873}             & \textbf{6.656} & \textbf{0.149} \\
\hline
\end{tabular}
}}
\vspace{1mm}
\caption{\color{black}\textbf{Make3D results with monocular training and $640\times192$ inputs.}
}
\label{Table:make3d}
\end{table}

\begin{table}[t]
\footnotesize
\centering
\resizebox{0.45\textwidth}{!}{
\setlength{\tabcolsep}{2.5mm}{
\begin{tabular}{c|c|c|c|c}
\bottomrule
 Architecture & Abs Rel & RMSE $\downarrow$ & FLOPs(B) & Params(M) \\
\hline
\hline
Monodepth2    & 0.115      &     4.863     & 8             & 14 \\
\hline
BRNet    & 0.105           &   4.462  & 31             & 19 \\
\hline
PackNet-SfM    & 0.111       &     4.601   & 205             & 128 \\
\hline
DIFFNet    & 0.102        &      4.483  & 2.3           & 12 \\
\hline
\rowcolor{green!10} DaCCN    & 0.099        &      4.316  & 4.3            & 13 \\
\hline
\end{tabular}
}}
\vspace{1mm}
\caption{\color{black}\textbf{Comparison results of params and computation.} }
\label{Table:flops}
\vspace{-2mm}
\end{table}

\section{Conclusion}
Monocular depth estimation is an ill-posed task and is very different from classification-based vision tasks in many ways. 
In this paper, we focus on the direction sensitivity and environmental dependency of this task and improve the efficiency of the backbone networks by exploiting a better feature representation. 
To achieve this, we propose a novel Direction-aware Cumulative Convolution Network (DaCCN) to strengthen the feature representation in two aspects: feature extraction and aggregation.
Firstly, the direction-aware module is developed to learn to adjust the feature extraction from each direction fully facilitating the encoding of different types of features.
Secondly, we propose the new cumulative convolution operation, which efficiently aggregates the information from the connection regions for improving environmental information aggregation. 
Experimental results show that the proposed models have achieved significant improvements on three prevalent benchmarks and set a new state-of-the-art performance.

{\small
\bibliographystyle{ieee_fullname}
\bibliography{egbib}
}

\end{document}